\documentclass{article}

\usepackage{PRIMEarxiv}

\usepackage[utf8]{inputenc} % allow utf-8 input
\usepackage[T1]{fontenc}    % use 8-bit T1 fonts
\usepackage{hyperref}       % hyperlinks
\usepackage{url}            % simple URL typesetting
\usepackage{booktabs}       % professional-quality tables
\usepackage{amsfonts}       % blackboard math symbols
\usepackage{nicefrac}       % compact symbols for 1/2, etc.
\usepackage{microtype}      % microtypography
\usepackage{lipsum}
\usepackage{fancyhdr}       % header
\usepackage{graphicx}       % graphics
\usepackage{enumitem}
\usepackage{array}
\usepackage{multirow}
\usepackage[table]{xcolor}
\usepackage{colortbl}
\usepackage{listings}
\usepackage{xcolor}
\graphicspath{{media/}}     % organize your images and other figures under media/ folder

\title{Bug in the Code Stack: Can LLMs Find Bugs in Large Python Code Stacks?
}

\author{
  Hokyung (Andy) Lee  \\
  Computer Science \\
  University of Waterloo \\
  \texttt{a362lee@uwaterloo.ca} \\
   \And
  Sumanyu Sharma \\
  Hamming.ai \\
  \texttt{sumanyu@hamming.ai} \\
  \And
   Bing Hu  \\
   Computer Science \\
  University of Waterloo \\
  \texttt{b25hu@uwaterloo.ca} \\
}

\newcommand{\graytext}[1]{\textcolor{gray}{#1}}

\definecolor{bg}{rgb}{0.98,0.98,0.98}
\definecolor{keyword}{rgb}{0.1,0.1,0.5}
\definecolor{comment}{rgb}{0.3,0.3,0.3}
\definecolor{string}{rgb}{0.4,0.0,0.0}

\usepackage{xcolor}

\definecolor{codegreen}{rgb}{0,0.6,0}
\definecolor{codegray}{rgb}{0.5,0.5,0.5}
\definecolor{codepurple}{rgb}{0.58,0,0.82}
\definecolor{backcolour}{rgb}{0.95,0.95,0.92}

\lstdefinestyle{mystyle}{
    backgroundcolor=\color{backcolour},   
    commentstyle=\color{codegreen},
    keywordstyle=\color{magenta},
    numberstyle=\tiny\color{codegray},
    stringstyle=\color{codepurple},
    basicstyle=\ttfamily\footnotesize,
    breakatwhitespace=false,         
    breaklines=true,                 
    captionpos=b,                    
    keepspaces=true,                 
    numbers=left,                    
    numbersep=5pt,                  
    showspaces=false,                
    showstringspaces=false,
    showtabs=false,                  
    tabsize=2
}

\begin{document}
\maketitle

\begin{abstract}
Recent research in Needle-in-a-Haystack (NIAH) benchmarks has explored the capabilities of Large Language Models (LLMs) in retrieving contextual information from large text documents. However, as LLMs become increasingly integrated into software development processes, it is crucial to evaluate their performance in code-based environments. As LLMs are further developed for program synthesis, we need to ensure that LLMs can understand syntax and write syntactically correct code. As a step in ensuring LLMs understand syntax, LLMs can be evaluated in their ability to find and detect syntax bugs.
Our benchmark, Bug In The Code Stack (BICS), is designed to assess the ability of LLMs to identify simple syntax bugs within large source code. Our findings reveal three key insights: (1) code-based environments pose significantly more challenge compared to text-based environments for retrieval tasks, (2) there is a substantial performance disparity among different models, and (3) there is a notable correlation between longer context lengths and performance degradation, though the extent of this degradation varies between models.
\end{abstract}

% keywords can be removed

\section{Introduction}
Recent advancements in Large Language Models (LLMs) have significantly increased their use in various real-world applications, including information retrieval and coding assistance \cite{github2021copilot}. Notably, the dramatic expansion of context window sizes in models like GPT-4 \cite{openai2024gpt4}, Claude 3 \cite{anthropic2024claude3}, and Gemini-1.5 \cite{geminiteam2024gemini} has broadened the potential applications of these models. To evaluate the retrieval capabilities of these LLMs within large context windows, a series of benchmarks known as Needle-in-a-Haystack (NIAH) \cite{gkamradt2023niah} has been developed.

The NIAH benchmarks \cite{gkamradt2023niah} typically involve prompting an LLM to retrieve contextual information based on a clue (e.g., needle) hidden within a large document (e.g., background). These benchmarks have been effective in evaluating LLMs' ability to retrieve information from large text data such as in text-summarization, and legal and medical domains \cite{kim2024fables,izzidien2024llm,saab2024capabilities}. NIAH represents important use-cases finding precedent case law in the legal domain \cite{izzidien2024llm} and information retrieval from lengthy electronic health records in the medical domain \cite{saab2024capabilities}. Verifying the "faithfulness" of long-text-summarization has also been shown as an important NIAH task for the FABLES dataset \cite{kim2024fables}.  

Generating code and programs following provided specifications or requirements is a long-standing challenge in computer science called program synthesis \cite{gulwani2017program}. LLMs have shown promise as tools for program synthesis with endeavours such as \textsc{CodeGen}\cite{nijkamp2023codegen}, \textsc{Codex}\cite{chen2021evaluating} powering Github Copilot \cite{github2021copilot}, and ChatGPT \cite{openai2024gpt4}. Equally as important is research evaluating code correctness and hallucinations when using LLMs as tools for program synthesis \cite{liu2023code, liu2024exploring}. Looking forward, LLMs can be powerful tools for information retrieval on large code-bases, such as in helping people with understanding code \cite{understandcode, liu2024repoqa}. However, there is a gap in research focusing on LLMs’ retrieval capabilities, particularly with the common problem of finding coding bugs, within large source code found in software development.

To address this research gap, we introduce Bug In The Code Stack (BICS), a benchmark specifically designed to assess LLMs’ capabilities in identifying bugs within large source code. In BICS, we automatically assemble Python source code from smaller code blocks to create the "background" and insert a syntactic bug within the source code as the "needle". With the assembled source code, we task the LLMs with identifying the line number and type of bug to evaluate the LLMs' debugging capabilities.

BICS benchmark makes several contributions to the NIAH research area:

\begin{itemize}[noitemsep,nolistsep]
    \item \textbf{Domain:} To our knowledge, BICS is the first benchmark aimed at evaluating LLMs' debugging capabilities in long-context source code.
    \item \textbf{Difficulty:} Our results demonstrate that BICS presents a significantly greater challenge to LLMs compared to existing NIAH benchmarks.
    \item \textbf{Relevancy:} BICS has been tested on all major models, providing a clear understanding of each model’s capabilities and informing best practices for their use as coding assistants in the industry.
\end{itemize}

\section{Related Work}

Previous studies have explored the capabilities of LLMs in various domains through several benchmarks designed to evaluate their capabilities at large context windows. For instance, RepoQA \cite{liu2024repoqa} benchmarks LLMs on long-context code understanding by evaluating their performance in retrieving specific functions from large codebases based on natural language descriptions. GILT is another tool to help developers understand their code and is an LLM information support tool directly builtin the IDE \cite{understandcode}. This work highlights the potential of LLMs in code search tasks, emphasizing the importance of understanding both the natural language descriptions and the code itself. 

Similarly, Kuratov et al. \cite{kuratov2024search} introduced BABILong, a benchmark designed to assess LLMs' performance in extracting and processing distributed facts within extensive texts. This benchmark demonstrates the ability of LLMs to handle extremely long input sequences, setting a new standard for evaluating the processing capabilities of LLMs for long contexts. LV-Eval \cite{yuan2024lveval} has been developed to standardize the NIAH benchmarks across bilingual datasets and multiple QA tasks. This benchmark consists of 11 datasets in English and Chinese applied to various QA tests, providing comprehensive bilingual measurements of LLMs' retrieval capabilities. RULER \cite{hsieh2024ruler} differentiates itself from standard NIAH benchmarks by introducing four task categories with thirteen representative tasks. These tasks include multi-hop tracing through variable tracking and aggregation through common/frequent word extraction, offering a broader assessment of LLMs' capabilities.

EvalPlus is a code synthesis evaluation framework to benchmark the functional correctness of LLM-generated code \cite{liu2023code}. It uses an LLM- and mutation-based strategy to automatically generate large amounts of test cases to test LLM-generated code compared to ground-truth implementations \cite{liu2023code}. Codex introduces HumanEval, an evaluation set that relies on manually constructed test cases to evaluate LLM solutions \cite{chen2021evaluating}. Many evaluation benchmarks, similar to EvalPlus and HumanEval, such as CoderEval \cite{Zhang_2024}, exist to evaluate the performance of the generated code. As LLMs are prone to generate hallucinations across applications, \textsc{HalluCode} is introduced to evaluate the performance of code LLMs in recognizing hallucinations that conflict with the user's requirements, contextual information, or code knowledge \cite{liu2024exploring}.

While these benchmarks offer valuable insights into the capabilities of LLMs with large context windows, they fail to address the specific needs of software engineering tasks, such as identifying bugs within large source code. Additionally, existing benchmarks often do not cover all the most popular models (e.g., GPT-4 \cite{openai2024gpt4}, Claude 3 \cite{anthropic2024claude3}, Gemini-1.5 \cite{geminiteam2024gemini}, Command-R \cite{cohere2024commandr}, Llama3 \cite{meta2024llama3}, etc.), making it difficult to evaluate the true capabilities of LLMs for real-world applications.

% \begin{listing}[h]
%     \centering
%     \begin{minted}{python}
%     1 | def fahrenheit_to_celsius(fahrenheit):
%     2 |   return (fahrenheit - 32) * 5.0/9.0
%     3 |
%     4 | def is_prime(num:
%     5 |     if num <= 1:
%     6 |         return False
%     7 |     for i in range(2, int(num**0.5) + 1):
%     8 |         if num % i == 0:
%     9 |             return False
%     10|     return True
%     Answer: 4, missing_parenthesis
%     \end{minted}
%     \caption{\textit{Example of syntactic bug identification within Python source code.}}
%     \label{lst:example_bug}
% \end{listing}

\section{Bug In The Code Stack}
\begin{table}[h]
    \centering
    \renewcommand{\arraystretch}{1.2} % Adjust the value as needed for more or less padding
    \setlength{\abovecaptionskip}{10pt} % Space above the caption
    % \rowcolors{1}{blue!10}{white} % Set the row colors
    % \resizebox{\textwidth}{!}{
    \begin{tabular}{>{\arraybackslash}m{0.3cm}|p{3.5cm}|p{11cm}}
    \hline
        \textbf{\#} & \textbf{Syntactic Bug} & \textbf{Description}\\
    \hline
        1 & Missing Colon & A colon is removed where previously used to
        indicate the start of a new block-of-code such as a loop, function or conditional statement. \\
        2 & Missing Parenthesis & A parenthesis is removed where previously one was used for the order of operations, order of evaluation, or in a tuple. \\
        3 & Missing Quotation & A quotation is removed where previously one was used to enclose a string. \\
        4 & Missing Comma & A comma is removed where previously one was used as a separator for items in a list, arguments, etc. \\
        5 & Mismatched Quotation & One of the quotations enclosing a string is replaced with a different quotation type, swapping between single quotation and double quotation. \\
        6 & Mismatched Bracket & One of the enclosing brackets for list, dictionary, function arguments, etc., is replaced with a different bracket type, swapping between parenthesis, square bracket, and curly bracket. \\
        7 & Keywords as Identifier & A variable name is set to a Python identifier name. \\
        \hline
    \end{tabular}
    % }
    \caption{Each of the 7 simple syntactical bugs.}
    \label{tab:listsyntax}
\end{table}
The Bug In The Code Stack (BICS) benchmark first aims to evaluate the ability of LLMs to find simple syntactical bugs in large code stacks. The intuition is to start with measuring LLMs for simple bugs before moving on to more complex ones. 
% \autoref{lst:example_bug} demonstrates an example syntactic bug of a missing parenthesis that is inserted into a code snippet.
BICS consists of insertions of 7 types of syntax bugs into a code stack.
\autoref{tab:listsyntax} lists and provides descriptions for each of the 7 syntactical bugs.
Bug In The Code Stack (BICS) benchmark consists of four major steps: Dataset Curation and Source Code Assembly, Bug Insertion, and Bug Retrieval. 

\subsection{Methodology}
% \begin{listing}[h]
%     \centering
    % \begin{minted}{python}
    \begin{lstlisting}[float,floatplacement=H, caption=Prompt template with two-shot examples for bug identification task., label=lst:twoshotexample]
    I will give you a codebase with a syntactic bug hidden at some line. You need to answer the line at which the syntactic error occurs and the type of the syntactic error.
    
    <example>
    1 | def fahrenheit_to_celsius(fahrenheit):
    2 |   return (fahrenheit - 32) * 5.0/9.0
    3 |
    4 | def is_prime(num:
    5 |     if num <= 1:
    6 |         return False
    7 |     for i in range(2, int(num**0.5) + 1):
    8 |         if num % i == 0:
    9 |             return False
    10|     return True
    Answer: 4, missing_parenthesis
    </example>
    
    <example>
    1| import random
    2| import string
    3|
    4| def generate_password(length=8):
    5|     characters = string.ascii_letters + string.digits
    6|     password = '\".join(random.choice(characters) for i in range(length))
    7|     return password
    Answer: 6, mismatched_quotation
    </example>
    
    <context>...</context>
    <bug_types>...</bug_types>
    
    Always return your answer in the following format: <line_number>, <bug_type>
    Do not write anything else after that.
    \end{lstlisting}
    % \caption{Prompt template with two-shot examples for bug identification task.}
    % \label{lst:twoshotexample}
% \end{listing}
% \textbf{Dataset Curation.} As a preliminary step to creating the benchmark, the Python Alpaca dataset \cite{bisht2024pythonalpaca} from HuggingFace was utilized to construct the "background" source code. The dataset's code samples, randomly distributed across various Python software domains (e.g., data structures and algorithms, machine learning), ensure that the "background" source code is not biased towards specific domains. Containing 18K Python code samples, the dataset was further filtered using a static code analyzer to remove samples with existing bugs and those deemed too small or large for the experiment.

\textbf{Code Curation.} We utilize a simple curation strategy to create the background code (i.e. haystacks) for our BICS benchmark. Our background source code is gathered from the Python Alpaca dataset \cite{bisht2024pythonalpaca}. The Python Alpaca dataset contains 18k Python code samples covering various software domains (i.e. data structures and algorithms, machine learning, databases, etc.). The Python Alpaca dataset is an unbiased code base from which we can create our haystacks. We remove any Alpaca code samples with existing syntax bugs with a static code analyzer to create a sanitized baseline for our experiments. Alpaca code samples deemed too small or large that fall outside the parameters of our experimentation are also removed. After removing code samples with pre-existing bugs, and ones that are too small or large, we are left with a filtered list of Alpaca code samples which we use to construct our haystack.

\textbf{Haystack (Code) Assembly.} We construct our haystack by stacking code samples from our filtered list of Alpaca code samples up to a desired token length. Desired lengths can be one of 500, 1k, 2k, 4k, 8k, or 16k tokens. Each sample in a stack is randomly selected from the filtered list of Alpaca code samples. As code samples that are too large or small are removed from our filtered list, each randomly selected Alpaca code sample is more consistent in size and uniformity is ensured comparing between haystacks. As pre-existing syntax bugs are removed from our filtered list, our haystacks are less noisy and we can accurately evaluate the ability of LLMs to detect simple bugs. The token length for our haystacks ranges between 500 to 16k tokens in each experiment. 

% The filtered code samples were then used to construct a source code up to the specified token length, with samples randomly selected and stacked until the desired length was reached. The prior filtering for consistent size and randomized selection ensures that the qualities of the "background" source code remain uniform across experiments. The target length for the "background" source code ranged from 500 tokens to 16K tokens in the experiments.

% \subsection{Bug Insertion}
\textbf{Needle (Syntax Bug) Insertion.} Once the Python source code haystack is assembled, a syntactic bug is inserted at 5 target depths ranging from 0\%, 25\%, 50\%, 75\%, and 100\% of the haystack. The benchmark employs seven different types of syntactic bugs, each capable of causing a runtime error. \autoref{tab:listsyntax} lists and provides descriptions for each of the 7 syntactical bugs.
Each syntactic bug has an equal chance of being selected for each experiment.

\textbf{Retrieving the (Bug) Needle.} After the bug is inserted into the Python source code, the LLMs are tasked with identifying the line number and type of bug within the code. The prompt template includes two-shot examples \cite{brown2020language} of short source code samples and a list of possible bug types to guide the LLMs. An example of a two-shot example pair is in listing \autoref{lst:twoshotexample}. 
The retrieval is considered a success if the LLM correctly finds both the line number and type of the inserted syntax bug.
This tasks evaluates the LLMs ability to understand code syntax as well as underscoring the importance of accurate retrieval in real-life software engineering scenarios.

\subsection{Experimental Setup}

\begin{table}[htbp]
    \centering
    \renewcommand{\arraystretch}{1.2} % Adjust the value as needed for more or less padding
    \setlength{\abovecaptionskip}{10pt} % Space above the caption
    % \rowcolors{1}{blue!10}{white} % Set the row colors
    % \resizebox{\textwidth}{!}{
    \begin{tabular}{>{\arraybackslash}m{0.5cm}|c|c|c|c|c}
    \hline
        \textbf{\#} & \textbf{Model Name} & \textbf{Publisher} & \textbf{Latest Release} & \textbf{Year Released} & \textbf{License}\\
    \hline
        1 & GPT-4o & OpenAI & 2024 & 2024 & Proprietary\\
        2 & GPT-4-Turbo & OpenAI & 2024 & 2023 & Proprietary\\
        3 & Claude-3.5 Sonnet & Anthropic & 2024 & 2024 &
        Proprietary\\
        4 & Claude-3 Opus & Anthropic & 2024 & 2024 & Proprietary\\
        5 & Gemini-1.5-Pro & Google & 2024 & 2024 & Proprietary\\
        6 & Gemini-1.5-Flash & Google & 2024 & 2024 & Proprietary\\
        7 & GPT-3.5-Turbo & OpenAI & 2024 & 2022 & Proprietary\\
        8 & Codestral & Mistral & 2024 & 2024 & Proprietary\\
        9 & Llama3-70b & Meta & 2024 & 2024 & Llama 3 Community\\
        10 & Command-r+ & Cohere & 2024 & 2024 & CC-BY-NC-4.0\\
        11 & Gemini-1.0-Pro & Google & 2024 & 2024 & Proprietary\\
        % 11 & CodeQwen-1.5-7b & Alibaba & 2023 & 2024 & Proprietary\\
        \hline
    \end{tabular}
    % }
    \caption{Each of the 11 LLMs that are tested against BICS.}
    \label{tab:listmodels}
\end{table}

\autoref{tab:listmodels} lists the 11 selected LLMs that are tested against the BICS benchmark. Given there are 6 options for haystack (code) token lengths (500, 1k, 2k, 4k, 8k, and 16k), and 5 options for needle (bug) depth (0\%, 25\%, 50\%, 75\%, and 100\%), each LLM is tested against the BICS benchmark with 30 combinations. Every combination is repeated 5 times so that each LLM is tested against the BICS benchmark for 150 iterations. As Llama3 \cite{meta2024llama3} has a maximum context window of up to 8k tokens, the 16k token length experiments are omitted. 

We utilized the LiteLLM \cite{berry2023litellm} package for inference. %, except for the CodeQwen1.5-7B-Chat model \cite{qwen2024codeqwen, qwen}, which was run locally.
To ensure consistency across model experiments, no modifications were made to the base prompt template to avoid skewing the experiment based on model-specific prompt optimizations. The only exceptions were Codestral \cite{mistral2024codestral} where we used Regex to filter out irrelevant trailing output common in Mistral models.

% The only exceptions were Codestral \cite{mistral2024codestral} 
% and CodeQwen1.5-7B-Chat: for Codestral, we used Regex to filter out irrelevant trailing output common in Mistral models, and for CodeQwen1.5-7B-Chat, we omitted two-shot examples, which often caused hallucinating answers in the smaller, 7B parameter model.

% We test 11 major models on the BICS benchmark, running each model 25 times for each combination of source code length (ranging from 500 to 16K tokens) and needle depth (ranging from 0\% to 100\%) and averaging the results across the 25 tests. The 16K context length experiments were omitted for the Llama3-70B model \cite{meta2024llama3}, which has a context window of up to only 8192 tokens. We utilized the LiteLLM \cite{berry2023litellm} package for inference, except for the CodeQwen1.5-7B-Chat model \cite{qwen2024codeqwen}, which was run locally. To ensure consistency across model experiments, no modifications were made to the base prompt template to avoid skewing the experiment based on model-specific prompt optimizations. The only exceptions were Codestral \cite{mistral2024codestral} and CodeQwen1.5-7B-Chat: for Codestral, we used Regex to filter out irrelevant trailing output common in Mistral models, and for CodeQwen1.5-7B-Chat, we omitted two-shot examples, which often caused hallucinating answers in the smaller, 7B parameter model.

\section{Results}

\begin{figure}[htbp]
    \centering
    \includegraphics[width=1.0\textwidth]{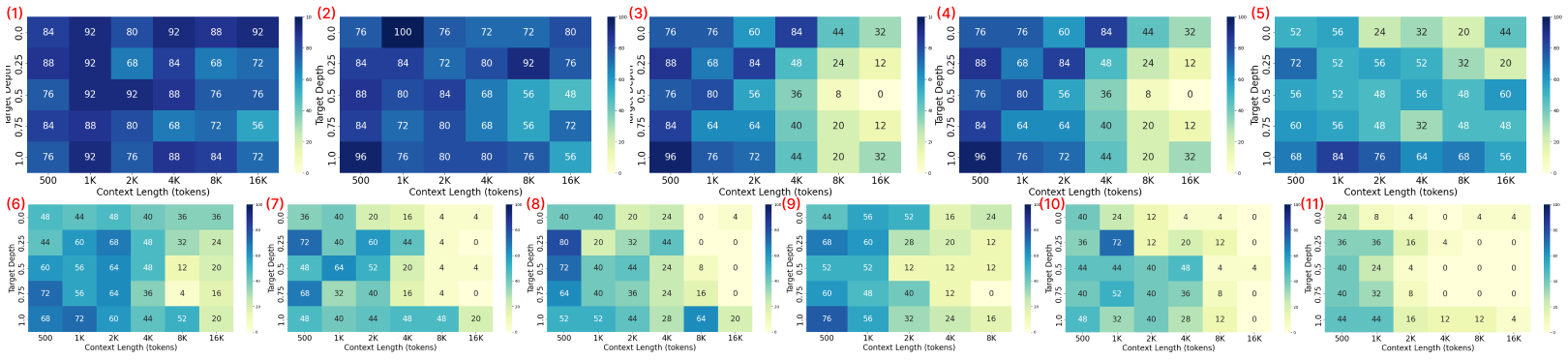}
    \caption{BICS results for all 11 selected models: (1) GPT-4o, (2) GPT-4-Turbo, (3) Claude-3.5 Sonnet, (4) Claude-3 Opus, (5) Gemini 1.5 Pro, (6) Gemini 1.5 Flash, (7) GPT-3.5 Turbo, (8) Codestral, (9) Llama3 70B, (10) Command-r+, and (11) Gemini 1.0 Pro.}
    \label{fig:overallresults}
\end{figure}

\autoref{fig:overallresults} contains the BICS benchmark for all 11 models for all 30 context length and needle length combinations, where the accuracy for each combination is averaged over 5 test iterations. 
\autoref{tab:model_performance} contains the performance at each context length for the 11 selected models. Accuracy results in \autoref{tab:model_performance} for each context length are generated by averaging results across needle depths for each context length. The average accuracy for all context lengths is reported alongside the standard deviation in \autoref{tab:model_performance}. Additionally, the average accuracy of each of the 11 models for all bug types is shown in \autoref{tab:bug_type_accuracy}.

% We include the results of 10 LLMs in Table 1. The performance at each context length is averaged across needle depths ranging from 0\% to 100\% (125 tests in total), while the last column represents the average accuracy across all tests (750 tests in total). Additionally, Figure 1 compares the variance in average accuracy for each needle depth across all models. Higher variance indicates greater sensitivity to needle placement, suggesting that the model is less robust. It is important to note that extremely low accuracy will result in low variance; thus, variance is more relevant when comparing models with similar performance levels.

\begin{table}[htbp]
    \centering
    \renewcommand{\arraystretch}{1.2} % Adjust the value as needed for more or less padding
    \setlength{\abovecaptionskip}{10pt} % Space above the caption
    % \rowcolors{1}{blue!10}{white} % Set the row colors
    % \resizebox{\textwidth}{!}{
        \begin{tabular}{>{\centering\arraybackslash}m{0.5cm}|l|c|c c c c c c|c}
            \hline
            \textbf{\#} & \textbf{Models} & \textbf{Ctx Size} & \textbf{500} & \textbf{1k} & \textbf{2k} & \textbf{4k} & \textbf{8k} & \textbf{16k} & \textbf{Avg.$\pm$(std)} \\
            \hline
            1 & gpt-4o & 128k & 82 & \textbf{91} & \textbf{79} & \textbf{84} & \textbf{78} & \textbf{74} & \textbf{81.2}$\pm$\graytext{(5.5)}\\
            2 & gpt-4-turbo & 128k & \textbf{86} & 82 & 78 & 74 & 70 & 66 & 76.1$\pm$\graytext{(6.7)}\\
            3 & claude-3.5-sonnet & 200k & 84 & 84 & \textbf{79} & 72 & 61 & 46 & 70.9$\pm$\graytext{(13.9)}\\
            4 & claude-3-opus & 200k & 84 & 73 & 67 & 50 & 23 & 18 & 52.5$\pm$\graytext{(24.8)}\\
            5 & gemini-1.5-pro & 1000k & 62 & 60 & 50 & 47 & 43 & 46 & 51.3$\pm$\graytext{(7.0)}\\
            6 & gemini-1.5-flash & 1000k & 58 & 58 & 61 & 43 & 27 & 23 & 45.1$\pm$\graytext{(15.2)}\\
            7 & gpt-3.5-turbo & 16k & 54 & 43 & 43 & 29 & 13 & 6 & 31.3$\pm$\graytext{(17.4)}\\
            8 & codestral & 32k & 62 & 38 & 35 & 29 & 18 & 5 & 31.1$\pm$\graytext{(17.7)}\\
            9 & llama3-70b & 8k & 60 & 54 & 33 & 17 & 13 & - & 35.4$^*$$\pm$\graytext{(19.1)}\\
            10 & command-r+ & 128k & 42 & 45 & 29 & 27 & 8 & 1 & 25.2$\pm$\graytext{(16.1)}\\
            11 & gemini-1.0-pro & 32k & 37 & 29 & 10 & 3 & 3 & 2 & 13.9$\pm$\graytext{(13.8)}\\
            % 11 & codeqwen1.5-7b-chat & 32k & 1 & 1 & 1 & 0 & 0 & 0 & 0.5 \graytext{(0.4)} \\
            \hline
        \end{tabular}
    % }
    \caption{Retrieval accuracy (\%) of 11 LLMs averaged over needle depth for each context length. The primary value in the "Avg." column represents the average accuracy across all context lengths. The standard deviation values are recorded in parenthesis in the "Avg." column. *Note that the average for Llama3-70b excludes the 16k context length combinations.}
    \label{tab:model_performance}
\end{table}

\begin{table}[htbp]
    \centering
    \renewcommand{\arraystretch}{1.2} % Adjust the value as needed for more or less padding
    \setlength{\abovecaptionskip}{10pt} % Space above the caption
    % \rowcolors{1}{blue!10}{white} % Set the row colors
    % \resizebox{\textwidth}{!}{
        \begin{tabular}{>{\centering\arraybackslash}l|c c c c c c c|c}
            \hline
            \textbf{Models} & \textbf{MC} & \textbf{MP} & \textbf{MQo} & \textbf{MCo} & \textbf{MQ} & \textbf{MB} & \textbf{KID} & \textbf{Avg.} \\
            \hline
             gpt-4o             & 93 & \textbf{93} & 18 & \textbf{90} & 91 & \textbf{96} & \textbf{80} & \textbf{81.2} \\
             gpt-4-turbo        & 94 & 92 & \textbf{59} & 78 & 69 & 73 & 65 & 76.1 \\
             claude-3.5-sonnet  & \textbf{96} & 76 &  5 & 74 & \textbf{92} & 65 & 76 & 70.9 \\
             claude-3-opus      & 76 & 76 & 53 & 59 & 28 & 23 & 46 & 52.5 \\
         gemini-1.5-pro         & 64 & 48 & 18 & 72 & 74 & 27 & 50 & 51.3 \\
             gemini-1.5-flash   & 81 & 73 & 32 & 58 & 14 & 11 & 41 & 45.1 \\
             gpt-3.5-turbo      & 35 & 65 & 18 & 44 & 24 & 21 & 5 & 31.3 \\
             codestral          & 45 & 41 & 36 & 50 & 20 & 5  & 16 & 31.1 \\
             llama-3-70b        & 47 & 52 & 26 & 28 & 37 & 23 & 33 & 35.4 \\
             command-r+         & 30 & 20 &  5 & 17 &  6 &  0 & 17 & 25.2 \\
             gemini-1.0-pro     & 17 &  0 &  6 & 30 & 17 & 20 & 5 & 13.9 \\
            \hline
             average & 65 & 61 & 26 & 54 & 43 & 32 & 42 & 43 \\
            \hline
        \end{tabular}
    % }
    \caption{Retrieval accuracy (\%) of each 7 syntactic bug types for each 11 LLMs. Each column name stands for the following: (MC) Missing Colon, (MP) Missing Parenthesis, (MQo) Missing Quotation, (MCo) Missing Comma, (MQ) Mismatched Quotation, (MB) Mismatched Bracket, (KID) Keywords as Identifier. The "Avg." column represents the average accuracy across all bug types within every context length.}
    \label{tab:bug_type_accuracy}
\end{table}

% \begin{figure}[htbp]
%     \centering
%     \includegraphics[width=0.75\textwidth]{./assets/variance_graph.png}
%     \caption{Variance in average accuracy for each needle depth.}
%     \label{fig:sample}
% \end{figure}

\section{Discussions}

\textbf{Disparity between Major Models.} The most notable observation is the performance disparity between major models. GPT-4o and GPT-4-Turbo exhibit the best performance among all the models. Claude 3 Opus, Gemini-1.5-Pro, and Gemini-1.5-Flash follow closely as the next best performers. Subsequently, GPT-3.5-Turbo, Codestral, Llama3-70B, and Command-R+ demonstrate similar performance levels. 
Lastly, Gemini-1.0-Pro shows lacklustre performance.

Comparing the standard deviation between model performances, we can see GPT-4o and GPT-4-Turbo has a high degree of performance while maintaining a low standard deviation. This shows that GPT-4 models are highly robust given this benchmark across all context lengths and needle depths. Higher variance indicates greater sensitivity to needle placement, suggesting that the model is less robust. It is important to note that extremely low accuracy will result in low variance; thus, variance is more relevant when comparing models with similar performance levels.

\textbf{Disparity between Retrieving Different Syntactical Bugs.} There is a noticeable difference in performance between different syntactical bugs. Generally speaking, the missing colon and missing parenthesis bugs were best retrieved compared to other bugs. The missing quotation and missing bracket bugs were very challenging to retrieve for most models. 

\textbf{Closed-Source Models vs. Open-Source Models.} With the exception of Gemini-1.0-Pro, all closed-source models demonstrate superior performance compared to open-source models, highlighting a gap in the development of current open-source models relative to their closed-source counterparts.

\textbf{Correlation between Context Length and Performance Degradation.} Generally, retrieval accuracy decreases as context length increases exponentially. However, some models are less affected by this increase in context length. For instance, GPT-4o's performance only drops by 8\% from 500 tokens to 16K tokens, and Gemini-1.5-Pro experiences a modest 16\% drop. In contrast, more susceptible models like Claude 3 Opus exhibit a dramatic 66\% decline in performance between 500 tokens and 16K tokens.

\textbf{Correlation between Needle Depth and Performance Degradation.} LLMs generally perform better when the syntax bug is placed at the code stack's top or bottom. There is a dip in performance for many models when the syntax bug is placed closer to the center of the code stack. Depending on the design of the LLM in question, syntax bugs nearer to the code stack's top or bottom may be easier to detect.

\begin{figure}[htbp]
    \centering
    \includegraphics[width=0.3\textwidth]{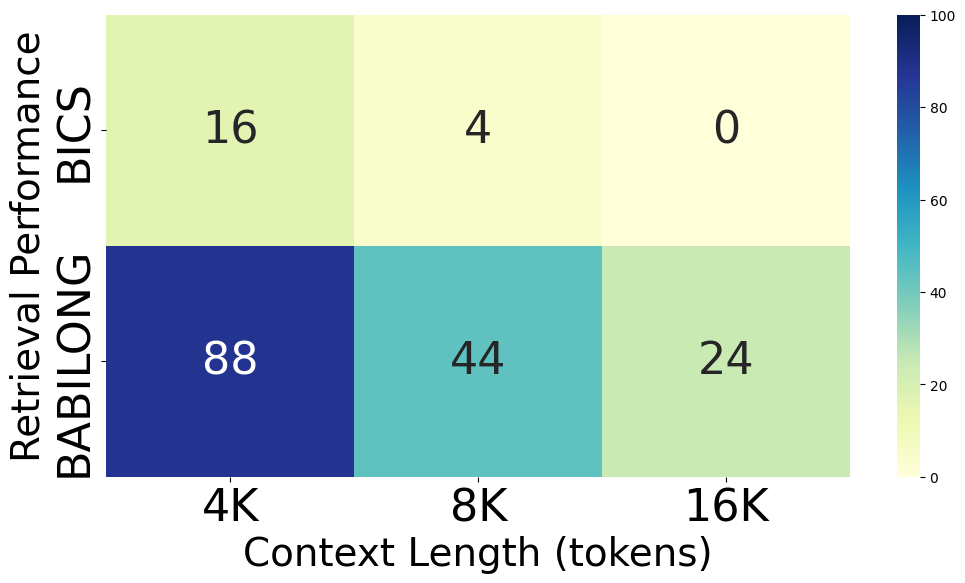}
    \caption{\textit{Comparing retrieval accuracy of GPT-3.5-Turbo on BICS versus BABILong benchmark}}
    \label{fig:sample}
\end{figure}

\textbf{Comparison with BABILong Benchmark.} Figure 2 compares the retrieval accuracy of GPT-3.5-Turbo on BICS with its performance on the BABILong benchmark \cite{kuratov2024search}. Notably, the same model exhibits 5-10 times lower accuracy on BICS, highlighting the increased difficulty of code-based benchmarks compared to text-based benchmarks for NIAH tests.

\section{Conclusion and Future Work}

We present Bug In The Code Stack (BICS), a benchmark designed to evaluate LLMs' capabilities in identifying bugs within large source code. By evaluating 11 major LLMs on the benchmark, we identified performance disparities between the models and observed the impact of increased context length on each model. Additionally, we established that code-based benchmarks pose significantly greater challenges compared to text-based benchmarks for NIAH tests, underscoring the need for further development to enhance model performance.

Our goal is to assist software developers in selecting the best models for developing their applications and to identify limitations in current models to guide further advancements. To this end, we plan to expand the benchmark to include runtime errors and warnings as the needle, as well as incorporate multilingual backgrounds for programming languages such as C++ and Rust. Specifically, we aim to address bugs related to pointers, memory management, and ownership \& borrowing in C++ and Rust, areas where, to our knowledge, a comprehensive LLM benchmark does not yet exist. 

\section*{Resource}

All datasets and evaluation codes are released at: \href{https://github.com/HammingHQ/bug-in-the-code-stack}{https://github.com/HammingHQ/bug-in-the-code-stack}

%Bibliography
\bibliographystyle{unsrt}
\bibliography{references}

\end{document}